# A Survey on Optical Character Recognition System

Noman Islam, Zeeshan Islam, Nazia Noor

*Abstract*—**Optical Character Recognition (OCR) has been a topic of interest for many years. It is defined as the process of digitizing a document image into its constituent characters. Despite decades of intense research, developing OCR with capabilities comparable to that of human still remains an open challenge. Due to this challenging nature, researchers from industry and academic circles have directed their attentions towards Optical Character Recognition. Over the last few years, the number of academic laboratories and companies involved in research on Character Recognition has increased dramatically. This research aims at summarizing the research so far done in the field of OCR. It provides an overview of different aspects of OCR and discusses corresponding proposals aimed at resolving issues of OCR.**

## I. INTRODUCTION

Optical Character Recognition (OCR) is a piece of software that converts printed text and images into digitized form such that it can be manipulated by machine. Unlike human brain which has the capability to very easily recognize the text/ characters from an image, machines are not intelligent enough to perceive the information available in image. Therefore, a large number of research efforts have been put forward that attempts to transform a document image to format understandable for machine.

OCR is a complex problem because of the variety of languages, fonts and styles in which text can be written, and the complex rules of languages etc. Hence, techniques from different disciplines of computer science (i.e. image processing, pattern classification and natural language processing etc. are employed to address different challenges. This paper introduces the reader to the problem. It enlightens the reader with the historical perspectives, applications, challenges and techniques of OCR.

## II. LITERATURE REVIEW

Character recognition is not a new problem but its roots can be traced back to systems before the inventions of computers. The earliest OCR systems were not computers but mechanical devices that were able to recognize characters, but very slow speed and low accuracy. In 1951, M. Sheppard invented a reading and robot GISMO that can be considered as the earliest work on modern OCR [1]. GISMO can read musical notations as well as words on a printed page one by one. However, it can only recognize 23 characters. The machine also has the capability to could copy a typewritten page. J. Rainbow, in 1954, devised a machine that can read uppercase typewritten English characters, one per minute. The early OCR systems were criticized due to errors and slow recognition speed. Hence, not much research efforts were put on the topic during 60's and 70's. The only developments were done on government agencies and large corporations like banks, newspapers and airlines etc.

Because of the complexities associated with recognition, it was felt that three should be standardized OCR fonts for easing the task of recognition for OCR. Hence, OCRA and OCRB were developed by ANSI and EMCA in 1970, that provided comparatively acceptable recognition rates[2] .

During the past thirty years, substantial research has been done on OCR. This has lead to the emergence of document image analysis (DIA), multi-lingual, handwritten and omni-font OCRs [2]. Despite these extensive research efforts, the machine's ability to reliably read text is still far below the human. Hence, current OCR research is being done on improving accuracy and speed of OCR for diverse style documents printed/ written in unconstrained environments. There has not been availability of any open source or commercial software available for complex languages like Urdu or Sindhi etc.

## III. TYPES OF OPTICAL CHARACTER RECOGNITION SYSTEMS

There has been multitude of directions in which research on OCR has been carried out during past years. This section discusses different types of OCR systems have emerged as a result of these researches. We can categorize these systems based on image acquisition mode, character connectivity, font-restrictions etc. Fig. 1 categorizes the character recognition system.

Based on the type of input, the OCR systems can be categorized as handwriting recognition and machine printed character recognition. The former is relatively simpler problem because characters are usually of uniform dimensions, and the positions of characters on the page can be predicted [3].

Handwriting character recognition is a very tough job due to different writing style of user as well as different pen movements by the user for the same character. These systems can be divided into two sub-categories i.e. on-line and off-line systems. The former is performed in real-time while the users are writing the character. They are less complex as they can capture the temporal or time based information i.e. speed, velocity, number of strokes made, direction of writing of strokes etc. In addition, there no need for thinning techniques as the trace of the pen is few pixels wide. The offline recognition systems operate on static data i.e. the input is a bitmap. Hence, it is very difficult to perform recognition.

There have been many online systems available because they are easier to develop, have good accuracy and can be incorporated for inputs in tablets and PDAs [4].

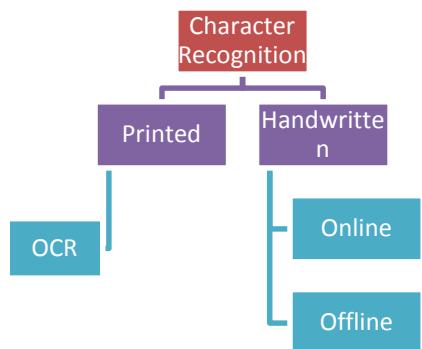

Figure.1: Types of character recognition system

## IV. APPLICATIONS OF OCR

OCR enables a large number of useful applications. During the early days, OCR has been used for mail sorting, bank cheque reading and signature verification [5]. Besides, OCR can be used by organizations for automated form processing in places where a huge number of data is available in printed form. Other uses of OCR include processing utility bills, passport validation, pen computing and automated number plate recognition etc [6]. Another useful application of OCR is helping blind and visually impaired people to read text [7].

## IV. MAJOR PHASES OF OCR

The process of OCR is a composite activity comprises different phases. These phases are as follows:

*Image acquisition:* To capture the image from an external source like scanner or a camera etc.

*Preprocessing:* Once the image has been acquired, different preprocessing steps can be performed to improve the quality of image. Among the different preprocessing techniques are noise removal, thresholding and extraction image base line etc.

*Character segmentation:* In this step, the characters in the image are separated such that they can be passed to recognition engine. Among the simplest techniques are connected component analysis and projection profiles can be used. However in complex situations, where the characters are overlapping /broken or some noise is present in the image. In these situations, advance character segmentation techniques are used.

*Feature extraction:* The segmented characters are then processes to extract different features. Based on these features, the characters are recognized. Different types of features that can be used extracted from images are moments etc. The extracted features should be efficiently computable, minimize intra-class variations and maximizes inter-class variations.

*Character classification:* This step maps the features of segmented image to different categories or classes. There are different types of character classification techniques. Structural classification techniques are based on features extracted from the structure of image and uses different decision rules to classify characters. Statistical pattern classification methods are based on probabilistic models and other statistical methods to classify the characters.

*Post processing:* After classification, the results are not 100% correct, especially for complex languages. Post processing techniques can be performed to improve the accuracy of OCR systems. These techniques utilizes natural language processing, geometric and linguistic context to correct errors in OCR results. For example, post processor can employ a spell checker and dictionary, probabilistic models like Markov chains and n-grams to improve the accuracy. The time and space complexity of a post processor should not be very high and the application of a post-processor should not engender new errors.

### a. Image Acquisition

Image acquisition is the initial step of OCR that comprises obtaining a digital image and converting it into suitable form that can be easily processed by computer. This can involve quantization as well as compression of image [8]. A special case of quantization is binarization that involves only two levels of image. In most of the cases, the binary image suffices to characterize the image. The compression itself can be lossy or loss-less. An overview of various image compression techniques have been provided in [9].

### b. Pre-processing

Next to image acquisition is pre-processing that aims to enhance the quality of image. One of the pre-processing techniques is thresholding that aims to binaries the image based on some threshold value [9]. The threshold value can be set at local or global level.

Different types of filters such as averaging, min and max filters can be applied. Alternatively, different morphological operations such as erosion, dilation, opening and closing can be performed.

Table.1: Major Phases of OCR system

| Phase | Description | Approaches |
|---|---|---|
| Acquisition | The process of acquiring image | Digitization, binarization, compression |
| Pre-processing | To enhance quality of image | Noise removal, Skew removal, thinning, morphological operations |
| Segmentation | To separate image into its constituent characters | Implicit Vs Explicit Segmentation |
| Feature Extraction | To extract features from image | Geometrical feature such as loops, corner points Statistical features such as moments |
| Classification | To categorize a character into its particular class | Neural Network, Bayesian, Nearest Neighborhood |
| Post-processing | To improve accuracy of OCR results | Contextual approaches, multiple classifiers, dictionary based approaches |

An important part of pre-processing is to find out the skew in the document. Different techniques for skew estimation includes: projection profiles, Hough transform, nearest neighborhood methods.

In some cases, thinning of the image is also performed before later phases are applied [10]. Finally, the text lines present in the document can also be found out as part of pre-processing phase. This can be done based on projections or clustering of the pixels.

### c. Character Segmentation

In this step, the image is segmented into characters before being passed to classification phase. The segmentation can be performed explicitly or implicitly as a byproduct of classification phase [11]. In addition, the other phases of OCR can help in providing contextual information useful for segmentation of image.

### d. Feature Extraction

In this stage, various features of characters are extracted. These features uniquely identify characters. The selection of the right features and the total number of features to be used is an important research question. Different types of features such as the image itself, geometrical features (loops, strokes) and statistical feature (moments) can be used. Finally, various techniques such as principal component analysis can be used to reduce the dimensionality of the image.

### e. Classification

It is defined as the process of classifying a character into its appropriate category. The structural approach to classification is based on relationships present in image components. The statistical approaches are based on use of a discriminate function to classify the image. Some of the statistical classification approaches are Bayesian classifier, decision tree classifier, neural network classifier, nearest neighborhood classifiers etc [12]. Finally, there are classifiers based on syntactic approach that assumes a grammatical approach to compose an image from its sub-constituents.

### f. Post-processing

Once the character has been classified, there are various approaches that can be used to improve the accuracy of OCR results. One of the approaches is to use more than one classifier for classification of image. The classifier can be used in cascading, parallel or hierarchical fashion. The results of the classifiers can then be combined using various approaches.

In order to improve OCR results, contextual analysis can also be performed. The geometrical and document context of the image can help in reducing the chances of errors. Lexical processing based on Markov models and dictionary can also help in improving the results of OCR [12].

## V. CONCLUSION

In this paper, an overview of various techniques of OCR has been presented. An OCR is not an atomic process but comprises various phases such as acquisition, pre-processing, segmentation, feature extraction, classification and post-processing. Each of the steps is discussed in detail in this paper. Using a combination of these techniques, an efficient OCR system can be developed as a future work. The OCR system can also be used in different practical applications such as number-plate recognition, smart libraries and various other real-time applications.

Despite of the significant amount of research in OCR, recognition of characters for language such as Arabic, Sindhi and Urdu still remains an open challenge. An overview of OCR techniques for these languages has been planned as a future work. Another important area of research is multi-lingual character recognition system. Finally, the employment of OCR systems in practical applications remains an active are of research.

## ACKNOWLEDGMENT

The authors would like to thank Iqra University, Karachi for their support in the completion of this research work.